\pdfoutput=1
\documentclass[letterpaper, 10 pt, conference]{ieeeconf}  

\IEEEoverridecommandlockouts                              

\usepackage{color}
\usepackage{xcolor}
\usepackage{xspace}
\usepackage{amsmath,amssymb}
\usepackage{mathrsfs}
\usepackage{graphicx}
\usepackage{subfigure}
\usepackage{booktabs}
\newcommand{\says}[3]{{\color{#3}#1: \emph{#2}\color{black}}\xspace}

\newcommand{\yc}[1]{\says{yc}{#1}{darkgreen}}
\newcommand{\zhcao}[1]{\says{zhcao}{#1}{brown}}
\newcommand{\method}{Safety Test framework by finding Av-Responsible Scenarios\xspace}
\newcommand{\methodabbr}{STARS\xspace}
\newcommand{\reward}{Hazard Arbitration Reward\xspace}
\newcommand{\rewardabbr}{HAR\xspace}
\newcommand{\scene}{AV-responsible scenarios\xspace}
\newcommand{\sceneabbr}{AV-responsible scenarios\xspace}

\newcommand{\signum}{\scriptsize}

\title{\LARGE \bf
Multi-Agent Vulnerability Discovery for Autonomous Driving\\with Hazard Arbitration Reward
}

\author{Weilin Liu$^{1*}$, Ye Mu$^{1*}$, Chao Yu$^{1}$, Xuefei Ning$^{1\dagger}$, \\
Zhong Cao$^{2}$, Yi Wu$^{3}$, Shuang Liang$^{4}$, Huazhong Yang$^{1}$ and Yu Wang$^{1\dagger}$ 
\thanks{$^{1}$W. Liu, Y. Mu, C. Yu, X. Ning, H. Yang, Y. Wang are with the Department of Electronic Engineering, Tsinghua University. }
\thanks{$^*$W. Liu, Y. Mu contribute equally to this work. }
\thanks{$^\dagger$Corresponding authors: Y. Wang {\tt\small yu-wang@tsinghua.edu.cn}, X. Ning {\tt\small foxdoraame@gmail.com}}%
\thanks{$^{2}$Z. Cao is with the School of Vehicle and Mobility, Tsinghua University}
\thanks{$^{3}$Y. Wu is with Shanghai Qi Zhi Institute and Institute for Interdisciplinary Information Sciences, Tsinghua University.}
\thanks{$^{4}$S. Liang is with Novauto Technology Co. Ltd, Beijing, China.}}

\begin{document}

\maketitle
\thispagestyle{empty}
\pagestyle{empty}

\begin{abstract}
Discovering hazardous scenarios is crucial in testing and further improving driving policies. However, conducting efficient driving policy testing faces two key challenges. On the one hand, the probability of naturally encountering hazardous scenarios is low when testing a well-trained autonomous driving strategy. Thus, discovering these scenarios by purely real-world road testing is extremely costly. On the other hand, a proper determination of accident responsibility is necessary for this task. Collecting scenarios with wrong-attributed responsibilities will lead to an overly conservative autonomous driving strategy. To be more specific, we aim to discover hazardous scenarios that are autonomous-vehicle responsible (AV-responsible), i.e., the vulnerabilities of the under-test driving policy. 

To this end, this work proposes a \method (\methodabbr) based on multi-agent reinforcement learning. \methodabbr guides other traffic participants to produce Av-Responsible Scenarios and make the under-test driving policy misbehave via introducing \reward (\rewardabbr). \rewardabbr enables our framework to discover diverse, complex, and AV-responsible hazardous scenarios. Experimental results against four different driving policies in three environments demonstrate that \methodabbr can effectively discover AV-responsible hazardous scenarios. These scenarios indeed correspond to the vulnerabilities of the under-test driving policies, thus are meaningful for their further improvements.
\end{abstract}

\section{INTRODUCTION}

\begin{figure*}[thpb]
      \centering
      \includegraphics[width=0.8\textwidth]{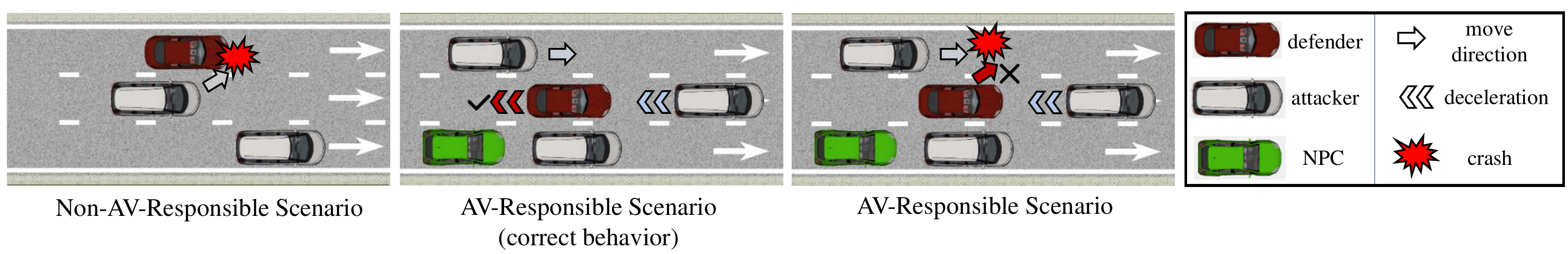}
      \caption{A comparative illustration of \scene. ``Defender'' denotes the vehicle controlled by the under-test policy. ``Attackers'' are the vehicles controlled by adversarial policies.
      ``NPC'' denotes other background vehicles. 
      Non-AV-responsible scenario: A scenario caused by other vehicles like unavoidable collision from a rushing attacker; \sceneabbr: A scenario caused by mistakes made by the under-test policy when a front car makes a slight deacceleration. Such accident corresponds to a vulnerability of the under-test policy.} 
      \label{fig:pc_scenes}
      \vspace{-3mm}
   \end{figure*}

Autonomous driving is a safety-critical application field, since autonomous vehicles directly interact with other vehicles and pedestrians, and the mistakes of the driving policy can result in severe accidents. Therefore, identifying possible hazardous scenarios of the autonomous driving policy through extensive testing is a key step in its iterative development process.
Among all possible accident scenarios, we focus on the Autonomous-Vehicle Responsible Scenarios (\scene) that correspond to the vulnerability of the driving policy. In other words, these accident scenarios are caused by the mistakes made by the driving policy.
We attribute the responsibility of the accident scenario according to the common sense of human judgment, excluding accident scenarios that are not caused or hard to avoid by the autonomous driving policy. To give an intuition on what an \sceneabbr hazardous scenario corresponding to a policy vulnerability is, we show a comparative illustration in Fig.~\ref{fig:pc_scenes}.

Since these hazardous cases are very sparse~\cite{kalra2016driving} and the road testing is costly, collecting a single AV-responsible accident case through real-world road testing can require thousands of miles of driving and lead to excessively high costs and unpredictable safety risks~\cite{Huang_2018_CVPR_Workshops,huang2019apolloscape}. 
Discovering \sceneabbr in simulators (e.g., CARLA~\cite{dosovitskiy2017carla}, SMARTS~\cite{zhou2020smarts}, AirSim~\cite{shah2018airsim}) is a promising way to mitigate the extremely high costs and safety risks of real-world testing. However, despite the high efficiency of these simulators in simulating driving experiences, it is still hard to find hazardous cases due to their rare occurrence.

In this paper, we propose the \method (\methodabbr), to efficiently discover vulnerabilities of an under-test driving policy. Specifically, we introduce Multi-Agent Reinforcement Learning (MARL) to control other traffic participants to interact with the under-test driving policy adversarially. In this way, the adversarial ``attackers'' can efficiently learn to construct hazardous scenarios where the under-test policy makes mistakes. Secondly, we present a novel reward design, \reward (\rewardabbr), which attributes the responsibility of hazardous scenarios according to common senses of human judgment. The power of \rewardabbr is three-fold: 1) It is agnostic to the concrete hazardous scenarios, thus enabling our framework to discover novel and diverse hazardous scenarios. 2) It prevents our framework from being stuck in trivial accident cases, and enables us to discover valuable \scene that indeed correspond to vulnerabilities of the under-test driving policy. 3) \rewardabbr is general across different environments.

We demonstrate the effectiveness of the proposed framework by testing against dynamic programming-based and reinforcement learning policies in different environments. Experimental results show that our framework can discover diverse and valuable vulnerabilities of the under-test policy: The ``attackers'' learn to collaboratively construct non-trivial hazardous scenarios in which the under-test policy makes mistakes. \methodabbr can be utilized as a pressure testing platform for driving policies or used to compare the robustness of policies. Also, it can be used to guide further design or learning of the driving policy.

In summary, the contributions of this paper are:

1) We propose a \method, which aims at finding essential \scene in the autonomous driving simulator.

2) We present a novel and general \rewardabbr design, and employ MARL to collaboratively construct complex \scene.

3) We experiment with four target driving polices, including dynamic programming based control policies and reinforcement learning policies in three different environments. The results demonstrate that \methodabbr can effectively discover diverse \scene.

\section{PRELIMINARY AND RELATED WORK}
\subsection{Partially Observed Markov Decision Process}
Our work can be describe as a Partially Observed Markov Decision Process (POMDP)~\cite{oliehoek2015concise}. A POMDP is defined by tuple $<\mathcal{S},\mathcal{A},\mathcal{O},\emph{N},\mathcal{P},\mathcal{R},\gamma>$. $\mathcal{S}$ presents state space including all possible conditions of \emph{N} agents. $\mathcal{A}$ is shared action space. $\mathcal{O}$ is observation space of each agent. $\mathcal{P}$ indicates the transition possibility model of states. $\mathcal{R}$ is the shared reward function. $\gamma\in[0,1]$ is the reward discount parameter. At each step, agent \emph{i} receives an observation $ \emph{$o_i$}=\mathcal{O}(\emph{$s_i$}) $ where $\emph{$s_i$}\in\mathcal{S}$ is the local state, takes an action \emph{$a_i$} according to shared policy $\pi_{\theta}(\emph{$a_i$}|\emph{$o_i$})$ where $\theta$ is the shared parameters, transit to next state $\emph{$s_i^\prime$}$ according to transition possibility $\emph{p}(\emph{$s_i^\prime$}|\emph{$s_i$},\emph{$a_i$})\in\mathcal{P}$, and finally receives a reward $\emph{$r_i$}=\mathcal{R}(\emph{$s_i$}|\emph{$a_i$})$. Considering homogeneous agents, each agent \emph{i} optimizes discounted cumulative rewards $\emph{$R_i$}=\sum_t\gamma^t\emph{$r_i^t$}$.

\subsection{Attacks on Policies}


Generally speaking, there are two types of strategies to attack decision-making policies: (1) \textbf{Attacking-on-States}: directly tampering with the state sequence of the target policy. These methods~\cite{lin2017tactics,lutjens2020certified} impose small disturbances on the state observations to make the policy output wrong decisions or get a smaller reward. (2) \textbf{Attacking-by-Policy}: learning one or more adversarial policies through interacting with the target policy. Gleave et al.~\cite{Gleave2020Adversarial} train an adversarial policy to compete with the under-test policy in a competitive game, and can discover flaws of the under-test policy.

We adopt the Attacking-by-Policy strategy in our work since it is more suitable for discovering vulnerabilities in autonomous driving policies. 
This is because that the perturbations crafted by Attacking-on-States strategies might be too noisy to occur in practice. In contrast, the constructed adversarial scenarios by Attacking-by-Policy strategies could appear naturally since the adversaries themselves are interactive agents with well-defined action spaces. 

There exist some studies~\cite{behzadan2019adversarial,feng2021intelligent,kuutti2020training,Wachi19} that adopt the Attack-by-Policy strategy to attack an autonomous driving policy. However, most of these studies aim at directly causing accidents~\cite{behzadan2019adversarial,feng2021intelligent,kuutti2020training}, rather than discovering the vulnerability of the under-test policy. We will elaborate on the multi-aspect differences between our work and these studies in Section~\ref{compare}.

\section{FRAMEWORK}


This section describes our framework that can efficiently discover Av-responsible hazardous scenarios to reveal the vulnerabilities of the driving policies. 
First, Section~\ref{sec:framework} gives out the illustration of the overall framework and introduces some terminologies.
To explore the hazardous scenarios efficiently, we employ MARL to train other adversarial agents, and this MARL framework is introduced in Section ~\ref{sec:mappo}. At the heart of our framework lies a novel reward design, \rewardabbr, which ensures that the discovered scenarios are Av-responsible and reveal vulnerabilities of the under-test policy. This reward design will be introduced in Section~\ref{sec:har}. Then, Section~\ref{sec:rw_compare} gives an in-depth comparison between related studies and our framework.

\begin{figure}[th]
      \centering
      \includegraphics[scale=0.4]{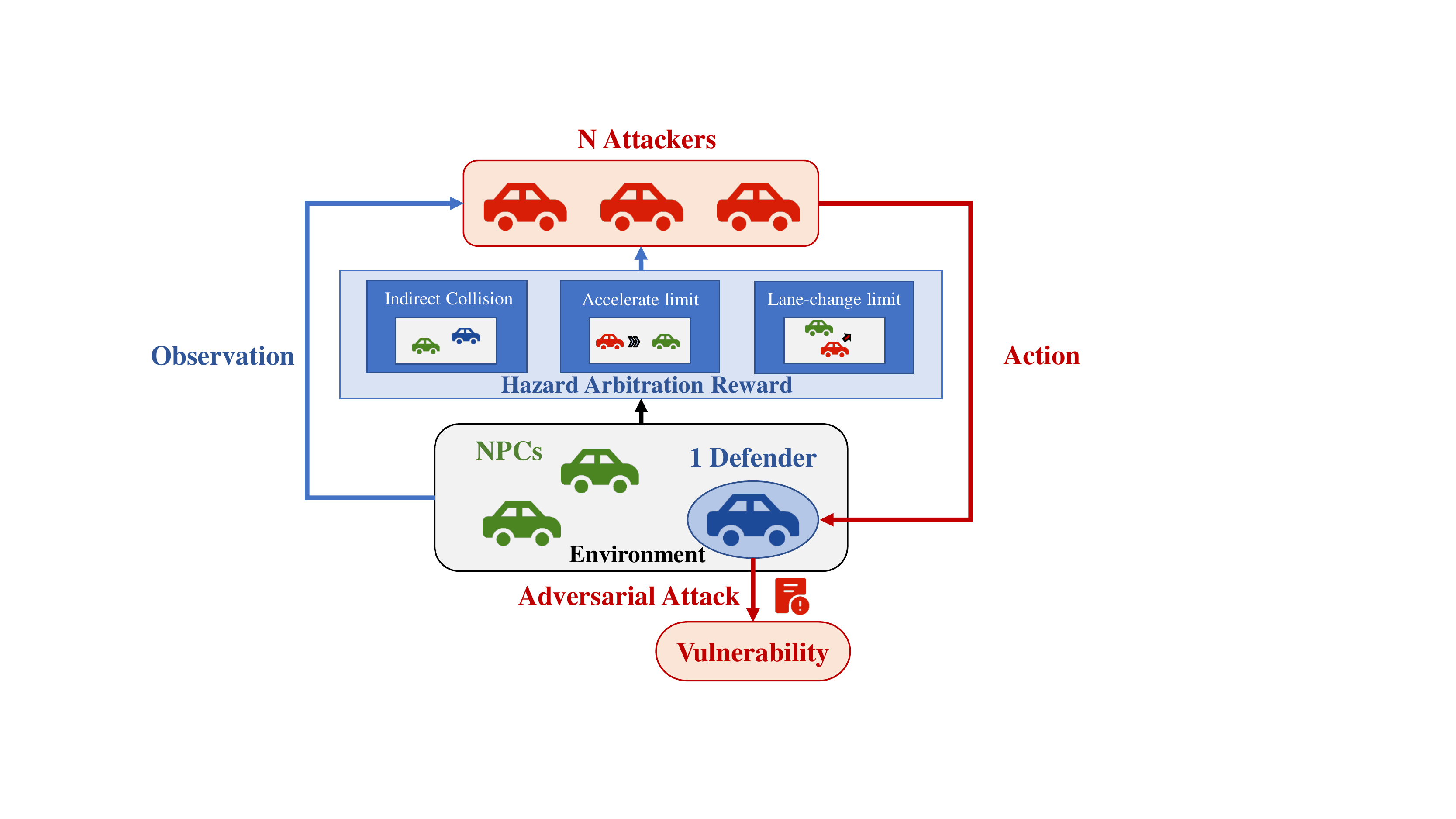}
      \caption{Overview of \methodabbr framework. Other traffic participants are controlled by adversarial policies learned by MAPPO, and the \rewardabbr reward design enables \methodabbr to discover diverse and valuable hazardous scenarios.} 
      \label{fig:framework}
\end{figure}

\subsection{\method}
\label{sec:framework}

We illustrate the overall framework of \method (\methodabbr) in Fig.~\ref{fig:framework}. We follow the terminologies used in the literature of adversarial attack methods, the under-test policy is defined as ``defender'', the adversarial vehicles trained by the MARL are called ``attackers'', and other background vehicles are named ``NPCs''. The defender, served as the attack target, is controlled by a well-trained or well-designed autonomous driving policy. The purpose of attackers is to explore the vulnerability that can make the defender behave wrongly, such as a car crash or out of bounds. In addition, we introduce NPCs to better simulate real driving scenarios.


\subsection{Multi-Agent PPO}
\label{sec:mappo}

We adopt the Multi-Agent PPO (MAPPO) framework~\cite{yu2021surprising} with parameter sharing. MAPPO is an extension of Proximal Policy Optimization (PPO)~\cite{schulman2017proximal} as an on-policy Centralized Training Decentralized Execution (CTDE) reinforcement learning algorithm. We train a actor network $\pi_\theta$ parameterized by $\theta$, and a value network $\emph{$V_\phi$}$ as the critic network. With Generalized Advantage Estimator method~\cite{schulman2015high}, we get Advantage function $\emph{$A_i^{(k)}$}$, and further the forms of actor and critic loss function are given as $$L(\theta)=[\frac{1}{Bn}\sum_{i=1}^B\sum_{k=1}^n min(r_{\theta,i}^{(k)}A_i^{(k)},clip(r_{\theta,i}^{(k)},1-\epsilon,1+\epsilon)A_i^{(k)})]$$
$$ +\sigma\frac{1}{Bn}\sum_{i=1}^B\sum_{k=1}^nS[\pi_\theta(o_i^{(k)})].$$
$$L(\phi)=\frac{1}{Bn}\sum_{i=1}^B\sum_{k=1}^n(max[(V_\phi(s_i^{(k)})-\hat{R}_i)^2,$$

$$ (clip(V_\phi(s_i^{(k)}),V_{\phi_{old}}(s_i^{(k)})-\varepsilon,V_{\phi_{old}}(s_i^{(k)})+\varepsilon)-\hat{R}_i)^2]),$$
where \emph{B} is the batch size, \emph{n} is the number of homogeneous agents, \emph{S} is the policy entropy, $\sigma$ is the entropy coefficient hyperparameter, and $\hat{R}_i$ is the discounted reward.

In our training environment, all agents share the same form of observation and action space. The $ith$ agent's observation $o_i=[k_0^i,k_1^i,k_2^i,k_3^i,k_4^i]$ describes the states of the closest five vehicles arranged by distance. $k=[p,x,y,v_x,v_y]$ is a state vector, where \emph{p} indicates presence of vehicle, \emph{x} and \emph{y} are x-coordinate and y-coordinate, and $v_x$ and $v_y$ are component velocities on x and y axis. The discrete action \emph{a} takes five choices for accelerating, decelerating, changing to left lane, changing to right lane, and doing nothing.


\subsection{\reward}
\label{sec:har}

Vulnerabilities of the driving policy cause the vehicle to take improper actions, such as unsafe lane changing or risky overtaking, which lead to accident scenarios.
However, some accident scenarios might not correspond to vulnerabilities in the under-test driving policy, since they are not caused or even not avoidable by the under-test driving policy, which means they are not valuable for improving the driving policy. 
An example of these non-AV-responsible hazardous scenarios is shown in Fig.~\ref{fig:pc_scenes} (Left), where the attacker suddenly makes a lane change and hits the under-test vehicle. In contrast, Fig.~\ref{fig:pc_scenes} (Right) shows a typical AV-responsible scenarios, where the under-test policy should have decelerated but makes a lane change and thus makes a collision.
To differentiate between AV-responsible and non-AV-responsible hazardous scenarios, we design a reward to penalize some trivial scenarios where the attackers are to blame. We name this reward design \rewardabbr, since this reward arbitrates the responsibility assignment of the accident scenes and provides behavior criteria for the attackers.


The design of \rewardabbr is as follows. The \rewardabbr is consisted of two parts: the collision reward \emph{$R_C$} and the aggressiveness penalty \emph{$P_{agg}$}. We give a sparse collision reward \emph{$R_C$} when the defender makes collisions with NPC vehicles or attacker vehicles, since we expect the attackers to explore how to generate the hazardous scenarios.Also we give an aggressiveness penalty \emph{$P_{agg}$} to limit the behaviors of the attackers.
Therefore, the \rewardabbr \emph{$R_{HAR}$} is
$$ R_C=
\begin{cases}
\phi& if\text{ the defender makes a collision} \\
0& otherwise
\end{cases}
$$
$$P_{agg}=
\begin{cases}
\rho& \left|\Dot{v_{att}} \right|>\lambda\; or \; a_{att}=a_{lc}\; or \; a_{att}=a_{rc} \\
0& otherwise
\end{cases}
$$
$$R_{HAR}=R_C + P_{agg},$$
where $\Dot{v_{att}}$ is the acceleration of the attacker, $\lambda$ is an acceleration threshold, $a_{att}$ is the action of the attacker, $a_{lc}$ and $a_{rc}$ are left lane change action and right lane change action.

\subsection{Comparison Between \methodabbr and Other Adversarial Attacks against Driving Policies}
\label{compare}
\label{sec:rw_compare}

\subsubsection{Single-Attacker Entrypoint}

There exist some studies that employ RL to learn one adversarial policy against the driving policy. 
Behzadan et al.~\cite{behzadan2019adversarial} apply a Deep Deterministic Policy Gradient (DDPG)~\cite{lillicrap2015continuous} to train the attacker vehicle in the urban road environment, whose goal is to make direct collision with the target vehicle.
Similarly, Feng et al.~\cite{feng2021intelligent} also set direct collisions as the goal of the attacker.
In \cite{kuutti2020training}, the authors put one Advantage Actor-Critic (A2C) attacker vehicle and one imitation learning-based defender vehicle on a one-lane one-way road. The defender is trained to stay at a safe distance and avoid collision with the leading attacker. The goal of the attacker is to make collisions, and the adversarial reward is the velocity of the defender divided by distance between two vehicles, limited to one hundred. This reward design is highly coupled with the simple car-following task, and is hard to generalize to other environments or tasks. 

Apart from the generalization ability of their method, three other issues also limit these methods' effectiveness in discovering the vulnerability of driving policies.

\begin{itemize}
    \item \textbf{Single-attacker entrypoint}: It is hard for these methods to discover complex hazardous scenarios, since their attacking entrypoint is a single vehicle, which is difficult to construct a complex hazardous scenario for the under-test policy.
    \item \textbf{Poor exploration}: The reward design in the literature is tightly coupled to concrete hazardous scenarios such as direct collision, which limits the exploration of novel hazardous scenarios.
    \item \textbf{No responsibility assignment}: The discovered scenarios are not valuable for improving the driving policy. A considerable part of direct collision scenarios do not correspond to the vulnerability of the driving policy, since these hazardous scenarios are not caused and even not avoidable by the under-test policy.
\end{itemize}

 In contrast, our work adopts a more powerful threat model, in which multiple adversarial vehicles are learned by MAPPO to construct more complex hazardous scenarios. Also, we develop \rewardabbr, a general reward design that enables our framework to explore novel, complex, and valuable \scene. In this way, \methodabbr discovers diverse vulnerabilities of the under-test policy, instead of constructing only a few hazardous scenarios that are not necessarily valuable for improving the driving policy.
 
\begin{figure*}[tb]
        \centering
        \subfigure[Highway]{
           \includegraphics[width=0.20\textwidth]{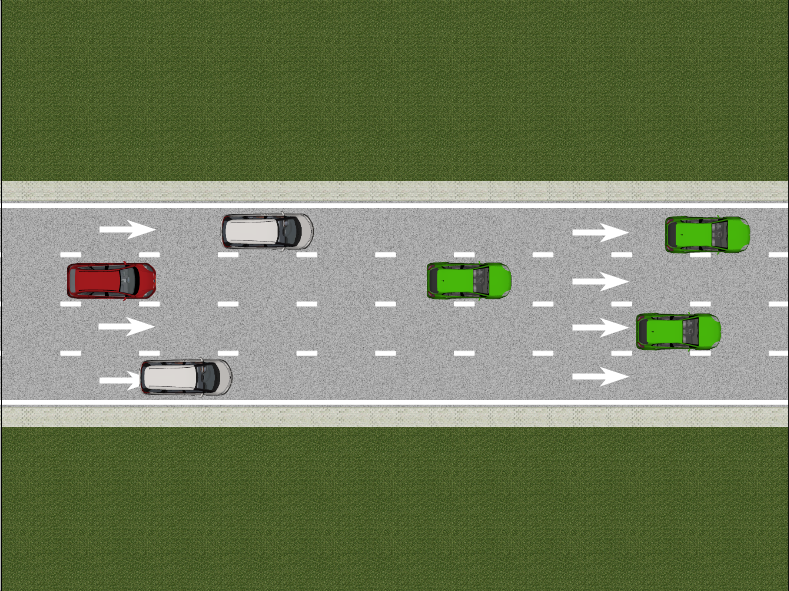}
           \label{Highway}
           }
        \subfigure[Merge]{
           \includegraphics[width=0.20\textwidth]{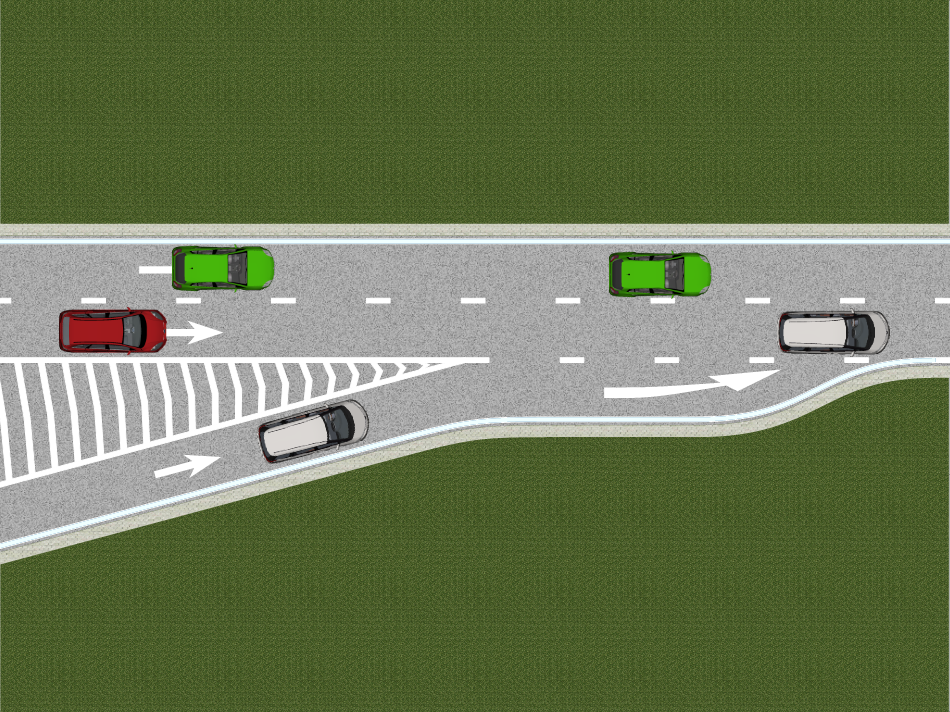}
           \label{Merge}
           }
        \subfigure[Roundabout]{
            \includegraphics[width=0.20\textwidth]{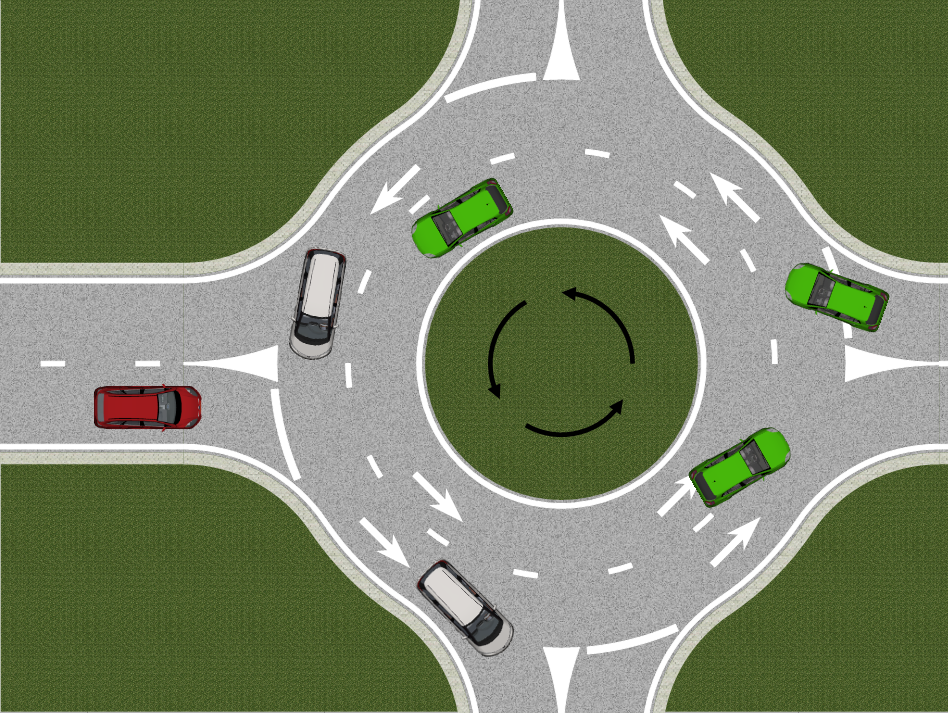}
            \label{Roundabout}
            }
        \centering
        \caption{Demonstration of the three driving environments. From left to right are ``Highway'', ``Merge'' and ``Roundabout''. The red vehicle is the ``defender''. The white vehicles are the ``attackers''. The green vehicles are the ``NPC'' vehicles.}
      \label{fig:highway}
    \end{figure*}


\subsubsection{Multi-Attacker Entrypoint}

Another relevant study~\cite{Wachi19} uses multiple adversarial vehicles, and designs a reward to encourage the attackers to reach their own goals as well as to attack the under-test policy. These two parts of reward design are named as the driving reward and adversarial reward, respectively.
However, this work fails to discover complex and diverse failure scenarios, since the design of driving reward is too restrictive for attackers to explore all interactive possibilities, and it also rules out many \scene. 
Moreover, they only conduct experiments against simple rule-based policies in a simple environment. With such a restrictive reward design and under such a simple setting, they fail to demonstrate any collaborative behavior of multiple attackers.

In contrast, \rewardabbr is a more suitable reward design for policy vulnerability discovery and guides our framework to discover diverse and complex \scene.  


\section{EXPERIMENT SETTING}

\subsection{Highway Environment Testbed}

To evaluate the proposed methods, this work chooses three typical driving environments, i.e., highway driving, merging, and roundabout driving. These three scenarios contain the most basic road elements like straight lane, merging lane, circle lane, and crossroad. Meanwhile, the combination of these scenarios could cover most safety-critical driving behaviors like lane changing, lane merging, overtaking, and crossroad driving \cite{kalra2016driving}. Our testbed is based on the previous work \cite{highway-env}, and we further
add three roles, i.e., attacker, defender, NPC. The details of these three scenarios are as follows:

\subsubsection{``Highway'' Scenario}
The ``Highway'' scenario defines a four-lane one-way road with several vehicles and infinite length. All NPC vehicles and the defender are initialized randomly on the road. The NPC vehicles closest to the defender are turned into attackers at the start of episodes. 

\subsubsection{``Merge'' Scenario}
The "Merge" scenario defines a two-lane one-way main road with one downside lane merging in. One of the attackers is set randomly on the downside lane. Other attackers, NPC vehicles, and the defender are initialized randomly on the main road.  

\subsubsection{``Roundabout'' Scenario}
The ``Roundabout'' scenario defines a circle roundabout two-lane road with two-lane entering straight roads from four directions. The defender is randomly set on the downside straight roads heading towards the circle road. The attackers and NPC vehicles are randomly initialized on the circle road.


\subsection{Tested Defender Policy for Autonomous Driving}
We evaluate four representative autonomous driving methods, which can be divided into 3 categories.
\begin{itemize}
    \item Planning-based algorithm: Value Iteration~\cite{bellman1966dynamic}
    \item Safe-enhanced algorithm: Robust Value Iteration~\cite{nilim2004robust}
    \item RL-based algorithm: Dueling Double Q-Network~\cite{wang2016dueling} and Policy Proximal Optimization~\cite{schulman2017proximal}
\end{itemize}
Value Iteration (VI) and Robust Value Iteration (RVI) are based on the finite MDP abstracted by the environment, and use dynamic-programming methods to optimize the Bellman Equation to get the action on each state. Dueling Double Deep Q-Network (D3QN) and Policy Proximal Optimization (PPO) defender models use the same observation and action space as the adversarial agent during the training process. Each model is trained in three scenarios respectively, aiming at collision avoidance and high-speed keeping. 

\subsection{Adversarial Settings For \scene}

We train MAPPO adversarial agents to explore the vulnerability of different driving policies. 
The reward function $\mathcal{R}$ of the attacker is consist of two parts: \rewardabbr $R_{\rewardabbr}$ and a distance reward $r_d$, i.e., $\mathcal{R}=R_{\rewardabbr}+r_d$. The \rewardabbr bounds the hazardous scenarios to encourage exploration of the \scene, with the parameter $\phi=10$ and $\rho=-10.5$. And we design a small distance reward $r_d$ to speed up our training.



\section{RESULTS AND DISCUSSION}
This section demonstrates the experimental results of the \methodabbr framework. First, we demonstrate and analyze the discovered AV-responsible scenes in Section~\ref{sec:exp_scenes}. Then, Section~\ref{sec:exp_singlemulti} compares the results of using single and multiple attackers. Finally, Section~\ref{sec:exp_compare} discusses the robustness comparison of four driving policies. 

\begin{figure*}[tb]
      \centering
      \includegraphics[width=0.8\textwidth]{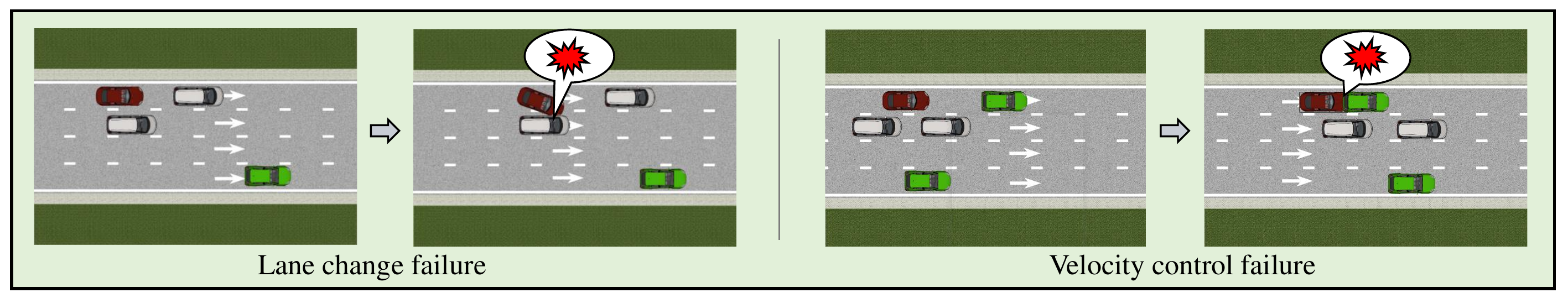}
      \caption{Diverse \scene of D3QN defender in Highway scenario}
      \label{fig:highway-diversity}
    \end{figure*}
    
   \begin{figure*}[tb]
      \centering
      \includegraphics[width=0.8\textwidth]{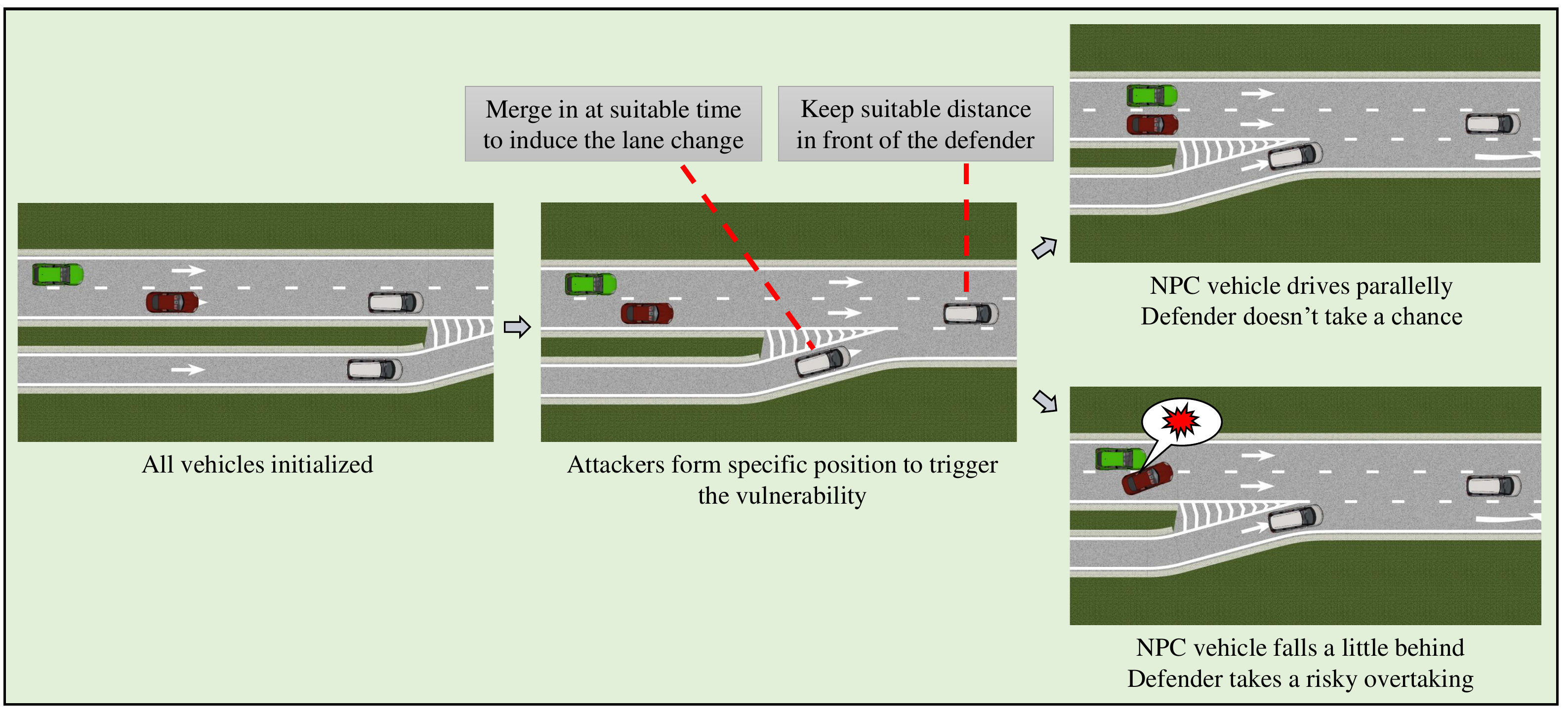}
      \caption{Collaborative attack example of D3QN defender in Merge scenario}
      \label{fig:merge-cooperation}
   \end{figure*}
   

\begin{table*}[tb]
\centering
\caption{Attack success rates (standard deviation) against four defender policies across three environments}
\label{tab:overall}
\begin{tabular}{cccccc}
\toprule
             & Scenarios   & VI                      & RVI                     & PPO                    & D3QN                   \\ 
\midrule
             & Highway    & 47.88\% \signum{(0.232)} & 47.06\% \signum{(0.143)} & 1.86\% \signum{(0.035)} & 13.64\% \signum{(0.115)} \\
1 attacker      & Merge      & 20.20\% \signum{(0.422)} & 32.30\% \signum{(0.818)} & 0   \% \signum{(0)} & 6.84 \% \signum{(0.018)} \\
             & Roundabout & 12.56\% \signum{(0.091)} & 13.11\% \signum{(0.093)} & 1.58\% \signum{(0.018)} & 8.55 \% \signum{(0.076)} \\ 
\midrule
             & Highway    & 76.08\% \signum{(0.214)} & 74.55\% \signum{(0.630)} & 2.32\% \signum{(0.060)} & 27.75\% \signum{(0.076)} \\
2 attackers     & Merge      & 77.18\% \signum{(0.922)} & 81.41\% \signum{(0.270)} & 3.23\% \signum{(0.130)} & 13.03\% \signum{(0.016)} \\
             & Roundabout & 19.78\% \signum{(0.122)} & 27.81\% \signum{(0.168)} & 9.66\% \signum{(0.079)} & 7.31 \% \signum{(0.133)} \\ 
\midrule
             & Highway    & 72.75\% \signum{(0.232)} & 85.49\% \signum{(0.110)} & 3.67\% \signum{(0.042)} & 58.24\% \signum{(0.178)} \\
3 attackers     & Merge      & 87.98\% \signum{(0.414)} & 90.33\% \signum{(0.092)} & 4.21\% \signum{(0.026)} & 30.99\% \signum{(0.016)} \\
             & Roundabout & 25.59\% \signum{(0.034)} & 27.00\% \signum{(0.294)} & 11.3\% \signum{(0.084)} & 20.75\% \signum{(0.225)} \\ 
\bottomrule
\end{tabular}
\vspace{-3mm}
\end{table*}

\subsection{The Discovered AV-Responsible Scenarios}
\label{sec:exp_scenes}

\methodabbr can discover AV-responsible hazardous scenarios that correspond to vulnerabilities of the under-test policy. Video demonstrations of the discovered scenarios are in the supplementary material. And we discuss three key characteristics of the discovered scenarios as follows.

\textbf{These scenarios are AV-responsible scenarios that correspond to vulnerabilities of the under-test policy.} We demonstrate three example hazardous scenarios discovered by \methodabbr in Fig.~\ref{fig:highway-diversity} and Fig.~\ref{fig:merge-cooperation}. It is obvious that in all these accidents, the under-test policy (red car) makes improper decisions and causes accidents, while other traffic participants do not exhibit any aggressive behavior. This indicates that the discovered hazardous scenarios are AV-responsible scenarios revealing vulnerabilities of the under-test policy. And the under-test policy should be improved to handle these scenarios.

\textbf{\methodabbr can discover diverse scenarios exhibiting various accident patterns,}
which is different from previous studies that conduct RL-based adversarial attacks on driving policies~\cite{behzadan2019adversarial,lillicrap2015continuous,feng2021intelligent,kuutti2020training,Wachi19}. Most previous studies~\cite{behzadan2019adversarial,lillicrap2015continuous,feng2021intelligent,kuutti2020training} aim to get direct collision scenarios by designing direct collision rewards. Consequently, their attackers can only learn to construct one type of direct collision scenarios. These scenarios are predictable before running the experiments thus only contribute minor knowledge to the policy testing process.

In contrast, we do not plant in the concrete hazardous scenario into the design of \rewardabbr, and the design of \rewardabbr enables the framework to explore and discover more diverse hazardous scenarios. Fig.~\ref{fig:highway-diversity} demonstrates two of our discovered hazardous scenarios in the Highway environment against the D3QN defender: 1) When the under-test vehicle is driving on the side lane, two attackers learn to form a siege by driving in front of and at the side of the under-test vehicle. The under-test D3QN policy will make a risky decision to overtake between the two attackers and cause a crash. 2) When the under-test vehicle is driving on the side lane following another vehicle (could be an NPC vehicle), two attackers can choose to occupy the adjacent lane. In this scenario, the under-test policy will fail to slow down and collide with the vehicle ahead.
Note that these two scenarios are constructed by the same attacker policy learned in a single experiment run, and the attackers can construct different scenarios according to different initialization of the vehicle positions.
This demonstrates the ability of \methodabbr to automatically discover diverse hazardous scenarios. 


\textbf{Multiple attackers exhibit collaborative behaviors to construct complex hazardous scenarios to make the under-test policy misbehave.}
Our application of MAPPO and design of \rewardabbr together enable the attackers to construct complex scenarios collaboratively. In another word, multiple attackers learn to collaborate to construct hazardous scenarios that cannot be constructed by a single attacker.

Fig.~\ref{fig:merge-cooperation} shows an example of the collaborative behavior of the attackers in the Merge environment. After initialization, the attacker on the main road keeps a suitable distance in front of the defender, while the attacker on the downside merging lane adjusts its velocity and merges in at the suitable time. This scenario will induce the under-test policy to make a lane change. At this time, if another vehicle is driving on the adjacent lane of the under-test vehicle and falls a little behind, the defender will fail to predict the collision and collide with it.
This scenario requires several attackers to appear at the right position and conduct the right action at the right time simultaneously, and it is extremely hard for a single attacker to trigger this scenario.

     \subsection{The Power of Using Multiple Attackers}
 \label{sec:exp_singlemulti}


To demonstrate the power of using multiple attackers, we compare the attack success rates of using single attacker and multiple attackers in Tab.~\ref{tab:overall}.
In the experiments of using single attacker, one of the attackers is changed into a non-adversarial NPC vehicle. The results show that the attack success rate of two or three attackers significantly outperforms that of single attacker.

\begin{figure}[tb]
  \centering
  \includegraphics[scale=0.4]{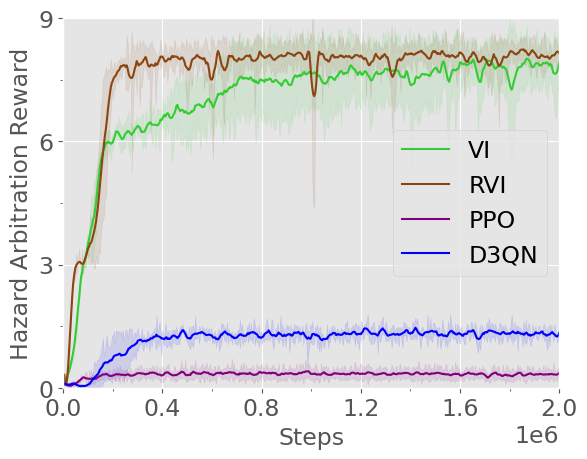}
  \caption{Training curves of \rewardabbr in the Merge scenario}
  \label{fig:Curves}
  \vspace{-3mm}
\end{figure}

\subsection{Robustness Comparison of Policies}
\label{sec:exp_compare}
We compare the robustness of four driving policies according to the attack success rates and the training curves. The attack success rates represent the occurrence rate of \scene, which indicates the risk factor of the under-test policy. The training curves in Fig.~\ref{fig:Curves} of \rewardabbr reflect the complexity and difficulty for attackers to explore. 

The hazardous scenarios exposing the vulnerability of the dynamic programming-based policies (VI and RVI) are easy to discover and highly reproducible. Limited by the finite estimation of infinite state-action pairs, many of their lane change decisions are risky. Thus, the attackers only need to drive close to the defender vehicle on the adjacent lanes to trigger the hazardous scenarios. Therefore, even one single attacker can successfully attack VI and RVI driving policies with high success rates.

Overall, the RL-based driving policies, PPO and D3QN, perform better than VI and RVI. Attacks against them have lower attack success rates in all environments. Discovering the vulnerability of the D3QN policy needs a sophisticated collaboration between attackers, as shown in Figure~\ref{fig:merge-cooperation}. PPO driving policy's vulnerability is even harder to discover, such that the single attacker fails to discover any vulnerability.

\section{CONCLUSION}

Discovering hazardous scenarios of the autonomous driving policy is a key step in its iterative development process.
In this paper, we propose the \methodabbr framework with a novel \rewardabbr design. Using MAPPO to adversarially control traffic participants and the design of \rewardabbr are the key points of our framework. They enable our framework to discover diverse, complex, and AV-responsible hazardous scenarios that are valuable for further improving the under-test driving policy.
Our framework is general w.r.t. different target driving policy and the driving environment. Experimental results in three typical driving environments (Highway, Merge, and Roundabout) against four driving policies (VI, RVI, D3QN, and PPO) demonstrate the effectiveness of our framework. Ablation studies demonstrate that our framework indeed takes advantage of the collaborative behaviors of the adversarial vehicles to construct complex hazardous scenarios. We also analyze the AV-responsible property and diversity of the discovered scenarios. Moreover, \methodabbr can be used to compare the robustness of driving policies.

We hope that our work can inspire further researches towards a pressure test platform for autonomous driving policies. Some interesting future directions include but are not limited to the following.
First, our reward design can be extended to incorporate more common senses or traffic rules~\cite{shalev2017formal,cao2020driving} according to actual needs. Second, considering the imperfectness of the perception and control modules in vulnerability discovery is worth investigating. Last but not least, combining \methodabbr and real-word testing (i.e., naturalistic driving) is an interesting direction. On the one hand, \methodabbr can play a role to efficiently augment the hazardous scenarios collected by real-world testing by using those scenarios as initialization states of the search process. On the other hand, one can efficiently construct and test the hazardous scenarios discovered by \methodabbr in the real world, saving thousands of miles of naturalistic driving.




\section*{ACKNOWLEDGMENT}
The authors gratefully acknowledge the support from TOYOTA. This work was also supported by Beijing National Research Center for Information Science and Technology (BNRist).

\bibliographystyle{IEEEtran}
\bibliography{reference}

\end{document}